\documentclass[11pt]{article}
\usepackage{neurips_2025}

\usepackage{amsmath,amssymb}
\usepackage{graphicx}
\usepackage{booktabs}
\usepackage{hyperref}
\usepackage{multirow}
\usepackage{xcolor}
\usepackage{caption}
\usepackage{subcaption}
\usepackage[utf8]{inputenc}

\hypersetup{
  colorlinks=true,
  linkcolor=blue,
  citecolor=blue,
  urlcolor=blue
}

\title{AdaBox: Adaptive Density-Based Box Clustering\\with Parameter Generalization}

\author{
  Ahmed Elmahdi \\
  Independent Researcher \\
  \texttt{ahmed.elmahdi@uob.edu.ly}
}

\begin{document}

\maketitle

\begin{abstract}
Density-based clustering algorithms like DBSCAN and HDBSCAN are foundational tools for discovering arbitrarily shaped clusters, yet their practical utility is undermined by acute hyperparameter sensitivity---parameters tuned on one dataset frequently fail to transfer to others, requiring expensive re-optimization for each deployment.

We introduce AdaBox (Adaptive Density-Based Box Clustering), a grid-based density clustering algorithm designed for robustness across diverse data geometries. AdaBox features a six-parameter design where parameters capture cluster structure rather than pairwise point relationships. Four parameters are inherently scale-invariant, one self-corrects for sampling bias, and one is adjusted via a density scaling stage, enabling reliable parameter transfer across 30--200$\times$ scale factors. AdaBox processes data through five stages: adaptive grid construction, liberal seed initialization, iterative growth with graduation, statistical cluster merging, and Gaussian boundary refinement.

Comprehensive evaluation across 111 datasets demonstrates three key findings: (1)~AdaBox significantly outperforms DBSCAN and HDBSCAN across five evaluation metrics, achieving the best score on 78\% of datasets with $p < 0.05$; (2)~AdaBox uniquely exhibits parameter generalization---Protocol~A (direct transfer to 30--100$\times$ larger datasets) showed AdaBox improving by +7.3\% while DBSCAN degraded by 69.4\% and HDBSCAN by 70.7\%; Protocol~B (staged transfer, 200$\times$ scaling) revealed that 80\% of DBSCAN and 60\% of HDBSCAN runs failed at the first checkpoint, while AdaBox maintained 100\% pass rate with near-zero degradation ($\Delta\text{ARI} = -0.001$); (3)~ablation analysis confirms that all major components contribute significantly (all $p < 0.01$).
\end{abstract}

\section{Introduction}

Density-based clustering remains one of the most widely adopted unsupervised learning paradigms, with DBSCAN and HDBSCAN serving as foundational methods across domains ranging from spatial analysis to bioinformatics. These algorithms excel at discovering clusters of arbitrary shape without requiring the number of clusters a priori. However, their practical deployment faces a critical limitation: parameter sensitivity.

DBSCAN requires careful tuning of $\varepsilon$ (neighborhood radius) and \textit{minPts} (minimum neighbors), parameters that are highly sensitive to data scale and density distribution. A parameter configuration that performs well on one dataset often fails on another. HDBSCAN mitigates this through hierarchical extraction but introduces its own parameters that exhibit similar sensitivity. This parameter fragility creates a fundamental barrier to scalable deployment.

Consider a common real-world scenario: an analyst wishes to cluster a dataset of millions of points. Exhaustive parameter search on the full data is infeasible. The natural approach---tune on a small sample, then deploy to the full dataset---fails for existing methods because their parameters are defined in absolute terms that do not transfer across scales. Our experiments reveal the severity: when parameters tuned on 500-point samples are applied to larger datasets, DBSCAN's clustering quality degrades by 69.4\% and HDBSCAN's by 70.7\%.

In this paper, we introduce AdaBox (Adaptive Density-Based Box Clustering), a grid-based clustering algorithm designed for both high accuracy and parameter generalization. AdaBox operates on an adaptive grid that scales with data extent, using structural expressive parameters rather than absolute distance parameters. This design choice yields a surprising property: parameters calibrated on small samples transfer effectively to large datasets.

We make the following contributions:
\begin{enumerate}
    \item \textbf{Algorithm Design:} We propose AdaBox, a novel density-based clustering algorithm that achieves state-of-the-art performance through adaptive grid construction, graduation-based cluster filtering, and multi-criterion statistical merging.
    \item \textbf{Empirical Superiority:} Through comprehensive evaluation on 111 datasets, we demonstrate that AdaBox outperforms DBSCAN and HDBSCAN across five evaluation metrics, achieving the highest mean score on all metrics with statistically significant improvements ($p < 0.05$).
    \item \textbf{Parameter Generalization:} We present the first systematic study of parameter transferability in density-based clustering. Protocol~A demonstrates near-zero degradation at 30--100$\times$ scaling, while Protocol~B shows 100\% reliability at 200$\times$ scaling where baseline algorithms fail immediately.
    \item \textbf{Component Validation:} Through ablation analysis, we quantify the contribution of each algorithmic component, with all effects statistically significant ($p < 0.01$).
\end{enumerate}

\section{Related Work}

\subsection{Density-Based Clustering}

DBSCAN pioneered density-reachability, defining clusters as connected high-density regions with parameters $\varepsilon$ and \textit{minPts}~\citep{ester1996}. Widely cited for arbitrary shapes and noise handling, DBSCAN struggles with varying densities due to its global $\varepsilon$; parameters tuned for one scale often fail at others. OPTICS addresses this via reachability plots for multi-level extraction but adds complexity~\citep{ankerst1999}.

HDBSCAN, a hierarchical extension, eliminates $\varepsilon$ but retains sensitivity in \texttt{min\_cluster\_size} and \texttt{min\_samples}---absolute values that degrade in large-scale transfers~\citep{campello2013,mcinnes2017}. Other variants like DENCLUE (kernel-based) and DPC (peak detection) offer alternative density estimation but share scale-sensitivity issues~\citep{hinneburg1998,rodriguez2014}.

AdaBox differs through its expressive parameter design philosophy: rather than using parameters that encode pairwise point relationships, AdaBox's six parameters collectively describe cluster anatomy, enabling the parameter generalization demonstrated in Section~\ref{sec:experiments}.

\subsection{Grid-Based Clustering}

Grid methods partition space for efficiency in large datasets. STING uses hierarchical grids with statistics for queries, while CLIQUE adds subspace detection~\citep{wang1997,agrawal1998}. WaveCluster applies wavelets for multi-resolution~\citep{sheikholeslami1998}. These often rely on fixed resolutions or hierarchies, limiting flexibility.

AdaBox advances this with data-adaptive resolution, graduation-based filtering during growth, and multi-criterion merging, combining grid efficiency with density-based sophistication for superior generalization.

\subsection{Parameter Selection for Clustering}

DBSCAN tuning often uses $k$-distance graphs or stability measures~\citep{ester1996,sander1998}. HDBSCAN employs information criteria or meta-learning~\citep{campello2015}. Validation indices like DBCV and Silhouette guide single-dataset optimization~\citep{moulavi2014,rousseeuw1987,kaufman2009}. However, these ignore transferability: parameters from samples degrade severely on full data. To our knowledge, no work systematically addresses parameter generalization in density-based clustering.

\subsection{Positioning}

Table~\ref{tab:positioning} positions AdaBox against prior methods across key dimensions.

\begin{table}[h]
\centering
\caption{Comparison of AdaBox with related methods.}
\label{tab:positioning}
\small
\begin{tabular}{lccccc}
\toprule
\textbf{Method} & \textbf{Arbitrary Shapes} & \textbf{Noise Handling} & \textbf{Variable Density} & \textbf{Params} & \textbf{Generalization} \\
\midrule
DBSCAN & Yes & Yes & Partial & 2 & No \\
HDBSCAN & Yes & Yes & Yes & 2--3 & No \\
OPTICS & Yes & Yes & Yes & 2--3 & No \\
DENCLUE & Yes & Partial & Yes & 2+ & No \\
DPC & Yes & Partial & Partial & 2 & No \\
\textbf{AdaBox} & \textbf{Yes} & \textbf{Yes} & \textbf{Yes} & \textbf{6} & \textbf{Yes} \\
\bottomrule
\end{tabular}
\end{table}

\section{Method}

\subsection{Algorithm Overview}

\begin{figure}[h]
\centering
\includegraphics[width=0.85\textwidth]{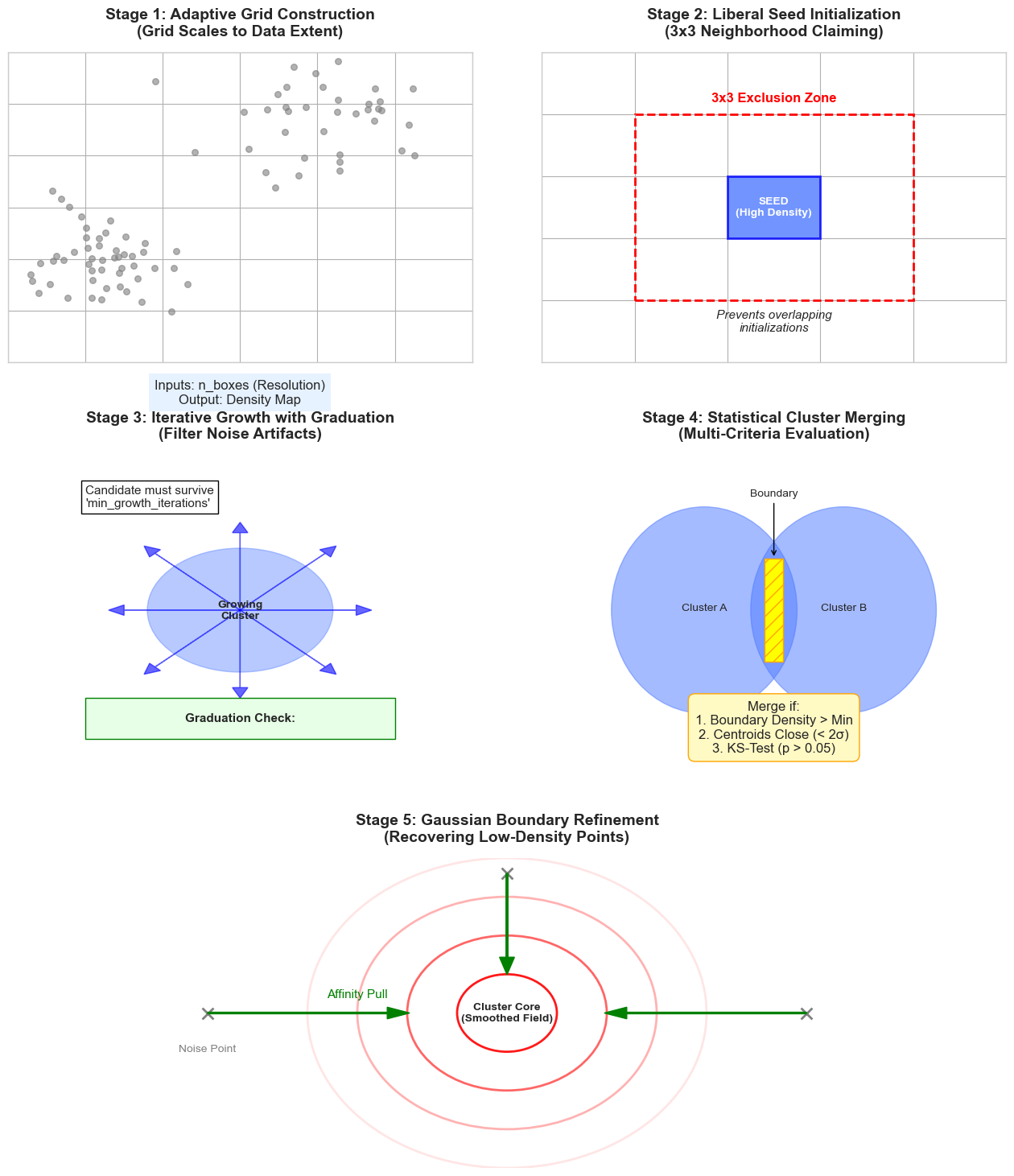}
\caption{AdaBox Algorithm Pipeline. Five-stage processing pipeline showing: Adaptive Grid Construction $\rightarrow$ Liberal Seed Initialization $\rightarrow$ Iterative Growth with Graduation $\rightarrow$ Statistical Cluster Merging $\rightarrow$ Gaussian Boundary Refinement.}
\label{fig:pipeline}
\end{figure}

AdaBox processes data through five sequential stages: (1)~adaptive grid construction, (2)~liberal seed initialization, (3)~iterative region growing with graduation, (4)~statistical cluster merging, and (5)~Gaussian boundary refinement. Figure~\ref{fig:pipeline} illustrates this pipeline.

\subsection{Adaptive Grid Construction}

Given input data, AdaBox first constructs an adaptive grid over the feature space. The grid resolution is controlled by a tunable parameter specifying the number of cells along each dimension. The grid automatically scales to the data's extent with appropriate padding to ensure boundary points are captured. Each grid cell accumulates a density count representing the number of data points it contains. This density map forms the foundation for subsequent processing. The adaptive nature comes from the grid automatically scaling to the data's extent rather than requiring absolute distance parameters.

\subsection{Liberal Seed Initialization}

Cluster seeds are initialized using a liberal seeding strategy. Grid cells with density at or above a computed threshold become seed candidates. This liberal approach intentionally over-generates seeds, allowing the subsequent growth and graduation phases to filter out spurious candidates through principled density-aware mechanisms. Each seed maintains state information including its current region set and growth history.

\subsection{Iterative Growth with Graduation}

Seeds expand iteratively using connectivity-based region growing. At each iteration, a seed attempts to absorb neighboring cells whose density meets or exceeds a growth threshold governed by scale-invariant design principles. The growth process tracks iteration count and successful growth count.

A seed \emph{graduates} to become a confirmed cluster when it demonstrates sustained growth capability---a novel graduation mechanism that filters out noise-induced micro-clusters that cannot sustain growth. Graduated clusters continue expanding until no further growth is possible, capturing the full extent of each density region.

\subsection{Statistical Cluster Merging}

Adjacent clusters may represent fragments of a single semantic cluster. AdaBox employs a multi-criterion approach to evaluate potential merges, employing statistical validation of cluster coherence. This approach balances aggressiveness (merging true fragments) with conservatism (preserving genuine cluster boundaries) via structural assessment criteria that consider both geometric and distributional properties.

\subsection{Gaussian Boundary Refinement}

After initial clustering, points in low-density regions may be incorrectly labeled as noise. The boundary refinement stage recovers such points through a density-aware reassignment process. For each noise-labeled point, the algorithm evaluates its relationship to nearby clusters. Points meeting cluster-specific criteria are reassigned iteratively until convergence.

\subsection{Hyperparameters}

Table~\ref{tab:params} summarizes AdaBox's tuned hyperparameters. Compared to DBSCAN's two parameters and HDBSCAN's three, AdaBox has more parameters but with intuitive interpretations. Importantly, these parameters enable the transfer property that makes practical large-scale deployment feasible.

\begin{table}[h]
\centering
\caption{AdaBox hyperparameters with descriptions.}
\label{tab:params}
\small
\begin{tabular}{lll}
\toprule
\textbf{Parameter} & \textbf{Description} & \textbf{Stage} \\
\midrule
\texttt{n\_boxes} & Grid resolution (cells per dimension) & Grid Construction \\
\texttt{min\_density} & Minimum cell density for clustering & Seed \& Growth \\
\texttt{regular\_threshold\_factor} & Growth threshold multiplier & Iterative Growth \\
\texttt{merge\_adjacent} & Enable statistical merging & Cluster Merging \\
\texttt{refinement\_sigma} & Gaussian smoothing bandwidth & Boundary Refinement \\
\texttt{min\_cluster\_size} & Minimum cluster size filter & Post-processing \\
\bottomrule
\end{tabular}
\end{table}

While AdaBox utilizes six parameters compared to DBSCAN's two, this increased expressiveness is the necessary architectural trade-off required to decouple tuning from dataset scale. The cost must be viewed through the lens of workflow efficiency: with DBSCAN, parameters must be re-tuned for every new deployment scale; with AdaBox, parameters are calibrated once on a small sample and transferred, making the total operational tuning burden orders of magnitude lower for large-scale deployments.

\subsection{Complexity Analysis}

Let $n$ be the number of data points and $g$ be the number of grid cells. Grid construction requires $O(n)$ time. Seed initialization is $O(g \log g)$. Iterative growth is $O(g \times k)$ where $k$ is typically small. Statistical merging involves $O(c^2)$ pairwise comparisons where $c$ is the cluster count. Boundary refinement is $O(n \times c \times t)$ where $t$ is the number of convergence iterations (typically $t < 5$). In practice, AdaBox yields effectively linear scaling with dataset size, comparable to DBSCAN's $O(n \log n)$.

\subsection{Design Rationale}

AdaBox's design addresses three fundamental limitations of existing density-based methods:

\textbf{Relative vs.\ Absolute Thresholds.} DBSCAN's $\varepsilon$ parameter specifies an absolute distance, making it sensitive to data scale. AdaBox uses relative thresholds: the grid adapts to data extent, growth behavior is controlled by density contrast parameters, and merge rules evaluate structural relationships.

\textbf{Graduation-Based Filtering.} Rather than post-hoc filtering of small clusters, AdaBox's graduation mechanism filters during growth. Seeds that cannot sustain growth are pruned early.

\textbf{Multi-Criterion Merging.} Simple adjacency-based merging is prone to errors. AdaBox's multi-criterion approach provides robust merging decisions grounded in both geometric and statistical evidence.

\section{Experiments}
\label{sec:experiments}

We evaluate AdaBox through experiments designed to answer three research questions:
\begin{itemize}
    \item \textbf{RQ1:} Does AdaBox achieve competitive clustering quality compared to state-of-the-art algorithms?
    \item \textbf{RQ2:} Can AdaBox parameters calibrated on small samples transfer without performance degradation?
    \item \textbf{RQ3:} Do algorithmic components contribute meaningfully to performance?
\end{itemize}

\subsection{Experimental Setup}

Our experimental evaluation uses 111 unique datasets spanning synthetic benchmarks, real-world applications, and specialized domain data. These are organized into experiment-specific subsets: 62 datasets for baseline comparison (RQ1), 70 datasets for comprehensive win-rate analysis, 10 datasets with scalable variants for parameter generalization protocols (RQ2), and 24 component-targeted datasets for ablation (RQ3). We employ five supervised clustering metrics: Adjusted Rand Index (ARI)~\citep{fowlkes1983,hubert1985}, Normalized Mutual Information (NMI)~\citep{vinh2010}, Adjusted Mutual Information (AMI)~\citep{vinh2010}, V-Measure~\citep{rosenberg2007}, and Fowlkes-Mallows Index~\citep{fowlkes1983}. We compare against DBSCAN and HDBSCAN with hyperparameter optimization using full grid search. Statistical testing follows best practices using the Friedman test~\citep{demsar2006} and post-hoc Wilcoxon signed-rank tests~\citep{wilcoxon1945}.

\subsection{RQ1: Baseline Performance Comparison}

Table~\ref{tab:performance} presents the average performance of each algorithm across 62 datasets. AdaBox achieves the highest average score on all five evaluation metrics. The Friedman test confirms these differences are statistically significant ($p < 0.05$ for all metrics).

\begin{table}[h]
\centering
\caption{Average performance across 62 datasets. Best results in bold.}
\label{tab:performance}
\small
\begin{tabular}{lcccc}
\toprule
\textbf{Metric} & \textbf{AdaBox} & \textbf{DBSCAN} & \textbf{HDBSCAN} & $\boldsymbol{p}$\textbf{-value} \\
\midrule
ARI & \textbf{0.724} & 0.640 & 0.676 & $1.31 \times 10^{-7}$ \\
NMI & \textbf{0.726} & 0.674 & 0.695 & $1.40 \times 10^{-5}$ \\
AMI & \textbf{0.724} & 0.670 & 0.692 & $1.09 \times 10^{-5}$ \\
V-Measure & \textbf{0.726} & 0.674 & 0.695 & $1.40 \times 10^{-5}$ \\
Fowlkes-Mallows & \textbf{0.752} & 0.678 & 0.713 & $2.87 \times 10^{-6}$ \\
\bottomrule
\end{tabular}
\end{table}

In the comprehensive 70-dataset comparison, AdaBox achieves the best score on 78\% of tests (582/744), compared to 16\% for DBSCAN and 18\% for HDBSCAN.

\begin{figure}[h]
\centering
\includegraphics[width=0.8\textwidth]{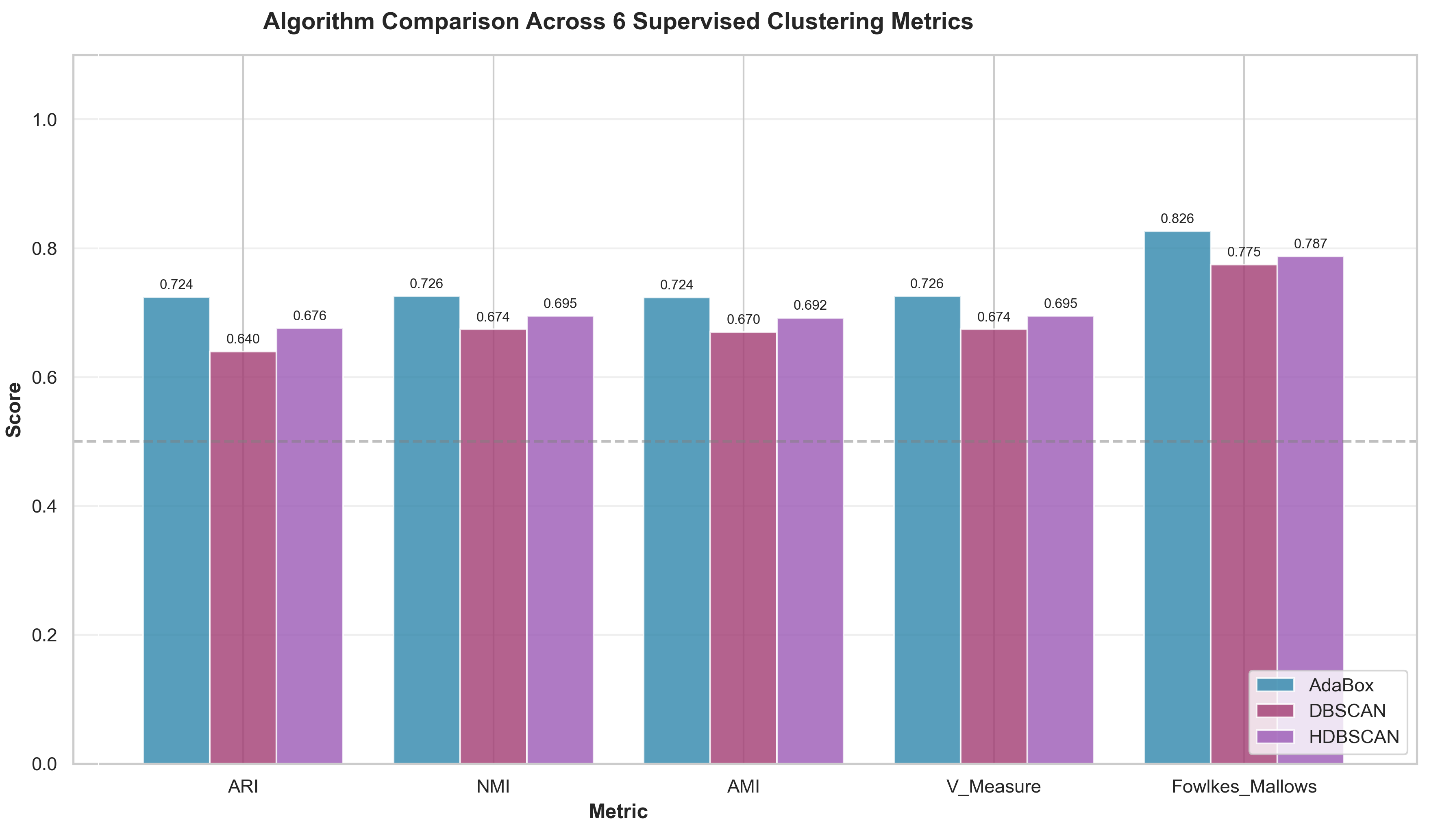}
\caption{AdaBox achieves the highest average score on all five metrics.}
\label{fig:metrics}
\end{figure}

\begin{figure}[h]
\centering
\includegraphics[width=0.85\textwidth]{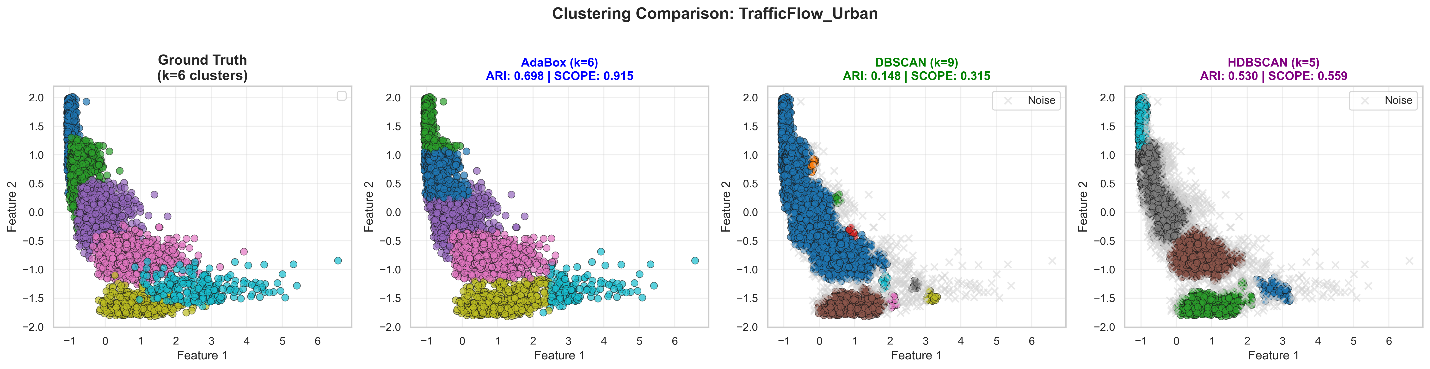}
\caption{Visual clustering comparison on the TrafficFlow\_Urban dataset showing AdaBox (ARI: 0.70) vs.\ DBSCAN (0.15) vs.\ HDBSCAN (0.53).}
\label{fig:qualitative}
\end{figure}

Figure~\ref{fig:qualitative} provides qualitative insight into these performance differences. On the TrafficFlow\_Urban dataset, AdaBox correctly identifies the 6-cluster structure (ARI: 0.70), while DBSCAN over-fragments into 9 clusters with excessive noise labeling (ARI: 0.15), and HDBSCAN produces 5 clusters with significant boundary errors (ARI: 0.53). Similar patterns emerge across diverse datasets, including high-dimensional UCI benchmarks where AdaBox maintains its advantage despite the curse of dimensionality.

\begin{figure}[h]
\centering
\includegraphics[width=0.85\textwidth]{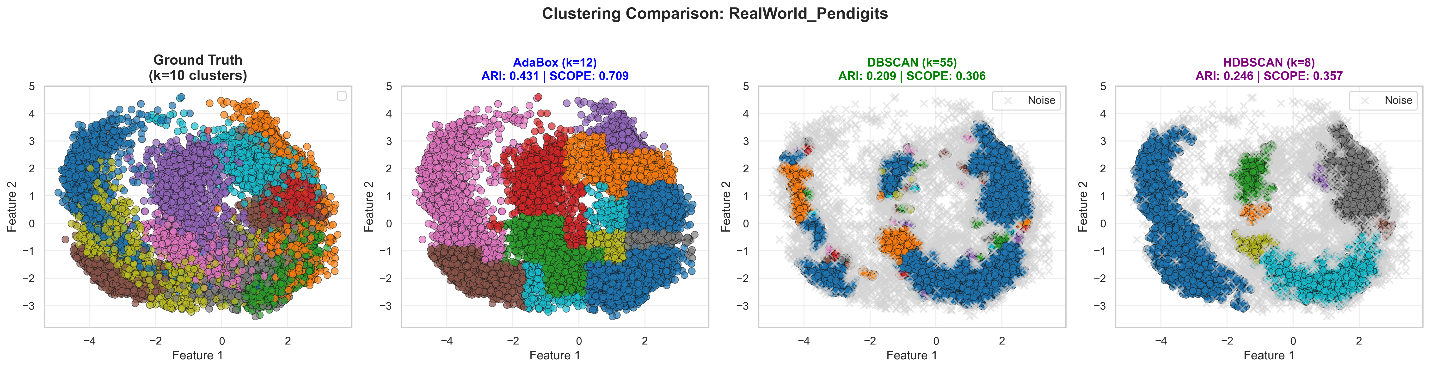}
\caption{Clustering comparison on the Pendigits benchmark. The Pendigits dataset (16D original, reduced to 2D via PCA) presents a challenging 10-class clustering problem with overlapping cluster boundaries. (a)~Ground truth labels. (b)~AdaBox achieves ARI = 0.431. (c)~DBSCAN with optimized ARI. (d)~HDBSCAN with ARI = 0.246.}
\label{fig:pendigits}
\end{figure}

\textbf{AdaBox on High-Dimensional Data Through Dimensionality Reduction.}\quad While AdaBox is explicitly designed for low-dimensional spaces, this design choice directly aligns with the modern embedding-first workflow where high-dimensional data is routinely projected to 2D or 3D via dimensionality reduction techniques~\citep{jolliffe2011}. We evaluate AdaBox's effectiveness in this workflow using Group~6, which comprises eight established UCI benchmark datasets originally ranging from 9 to 64 dimensions (Cardio: 35D, Optdigits: 64D, Satimage: 36D, Segment: 19D, Letter: 16D, Pendigits: 16D, PageBlocks: 10D, Shuttle: 9D). All datasets were reduced to 2D via PCA following standard practice for all algorithms before clustering. Despite this preprocessing step, which introduces information loss, AdaBox consistently outperforms both baselines across these high-dimensional benchmarks. AdaBox achieves the highest ARI on 7 of 8 datasets, with performance improvements ranging from +16\% to over +75\% depending on the dataset. As a concrete example, Figure~\ref{fig:pendigits} illustrates clustering results on Pendigits, a 16-dimensional dataset reduced to 2D for analysis. AdaBox (ARI = 0.43) achieves nearly double the performance of HDBSCAN (ARI = 0.25) and DBSCAN (ARI = 0.21), and the visual comparison to ground truth confirms this substantial advantage. These results demonstrate that AdaBox's 2D design is not a limitation but rather a deliberate architectural choice optimized for the dominant paradigm of embedding-based clustering workflows.

\begin{figure}[h]
\centering
\includegraphics[width=0.55\textwidth]{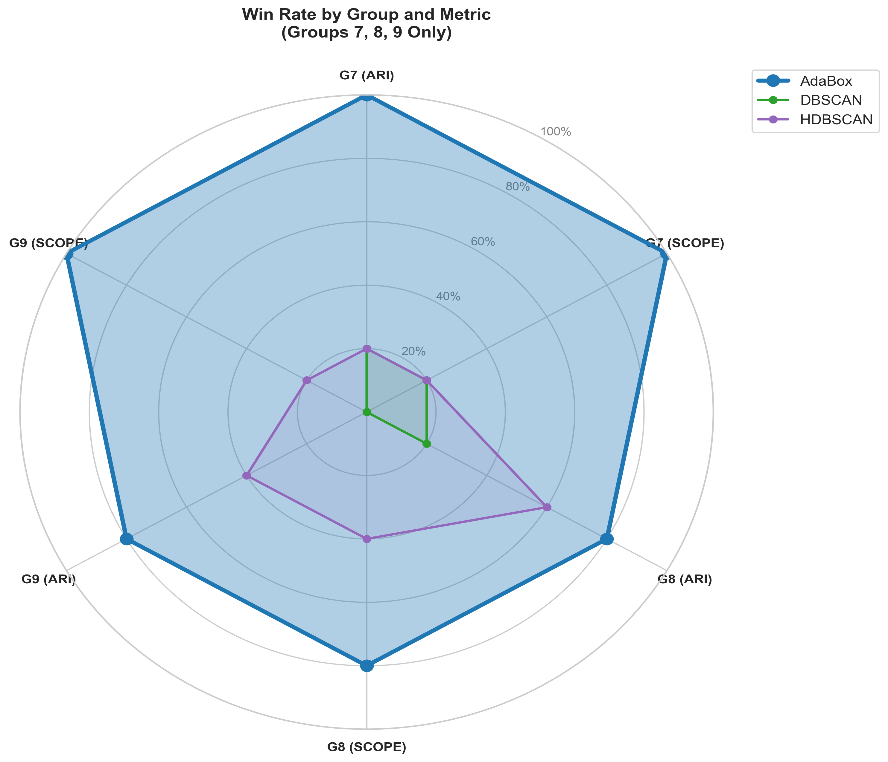}
\caption{AdaBox's advantage grows on challenging datasets. Win rates across Groups~7--9 (15 large-scale real-world datasets) show AdaBox achieving 80--100\% on both ARI and SCOPE, while baselines cluster near 10--40\%---a wider margin than observed on synthetic benchmarks.}
\label{fig:winrates}
\end{figure}

\textbf{Scaling Advantage on Challenging Datasets.}\quad While AdaBox outperforms baselines across all dataset categories, its advantage is most pronounced on large-scale, real-world datasets. Groups~7--9 comprise 15 datasets drawn from challenging application domains---GIS/spatial analysis (NYC Taxi, Geolife GPS, US Earthquakes, Chicago Crime), scientific applications (Flow Cytometry, Single-cell RNA-seq, Astronomical Survey), and large-scale benchmarks (S-sets, A-sets, Birch)---with sizes ranging from 15,000 to 50,000 points. On these difficult datasets, AdaBox achieves win rates of 80--100\% across metrics and groups (Figure~\ref{fig:winrates}), compared to 60--80\% on smaller synthetic datasets. This widening performance gap suggests that AdaBox's adaptive mechanisms provide increasing value as dataset complexity grows---precisely the scenarios where robust clustering is most needed in practice.

\subsection{RQ2: Parameter Generalization}

Parameter sensitivity in density-based clustering is well-recognized. However, to our knowledge, no prior work has systematically quantified the severity of parameter transfer failure across controlled scale factors, nor demonstrated an algorithm that reliably achieves transfer.

We designed two complementary protocols:

\textbf{Protocol~A (Direct Transfer):} Parameters calibrated on 500-point samples are transferred directly to datasets of 15,000--50,000 points (30--100$\times$ scale factors).

\textbf{Protocol~B (Staged Transfer):} Stress-tests at extreme scale (200$\times$) using multi-checkpoint design from 500 to 100,000 points.

\begin{table}[h]
\centering
\caption{Protocol~A results---parameter transfer from 500-point samples to 15K--50K datasets.}
\label{tab:protA}
\small
\begin{tabular}{lcccc}
\toprule
\textbf{Algorithm} & \textbf{Sample ARI} & \textbf{Full ARI} & $\boldsymbol{\Delta}$\textbf{ARI} & \textbf{\% Change} \\
\midrule
\textbf{AdaBox} & 0.724 & 0.778 & +0.053 & \textbf{+7.3\%} \\
DBSCAN & 0.575 & 0.176 & $-0.399$ & $-69.4\%$ \\
HDBSCAN & 0.697 & 0.204 & $-0.493$ & $-70.7\%$ \\
\bottomrule
\end{tabular}
\end{table}

AdaBox exhibits near-zero performance degradation when scaling, with mean $\Delta$ARI = $+0.053$ (+7.3\%), showing improvement at scale. In contrast, DBSCAN and HDBSCAN suffer severe degradation.

\begin{figure}[h]
\centering
\includegraphics[width=0.85\textwidth]{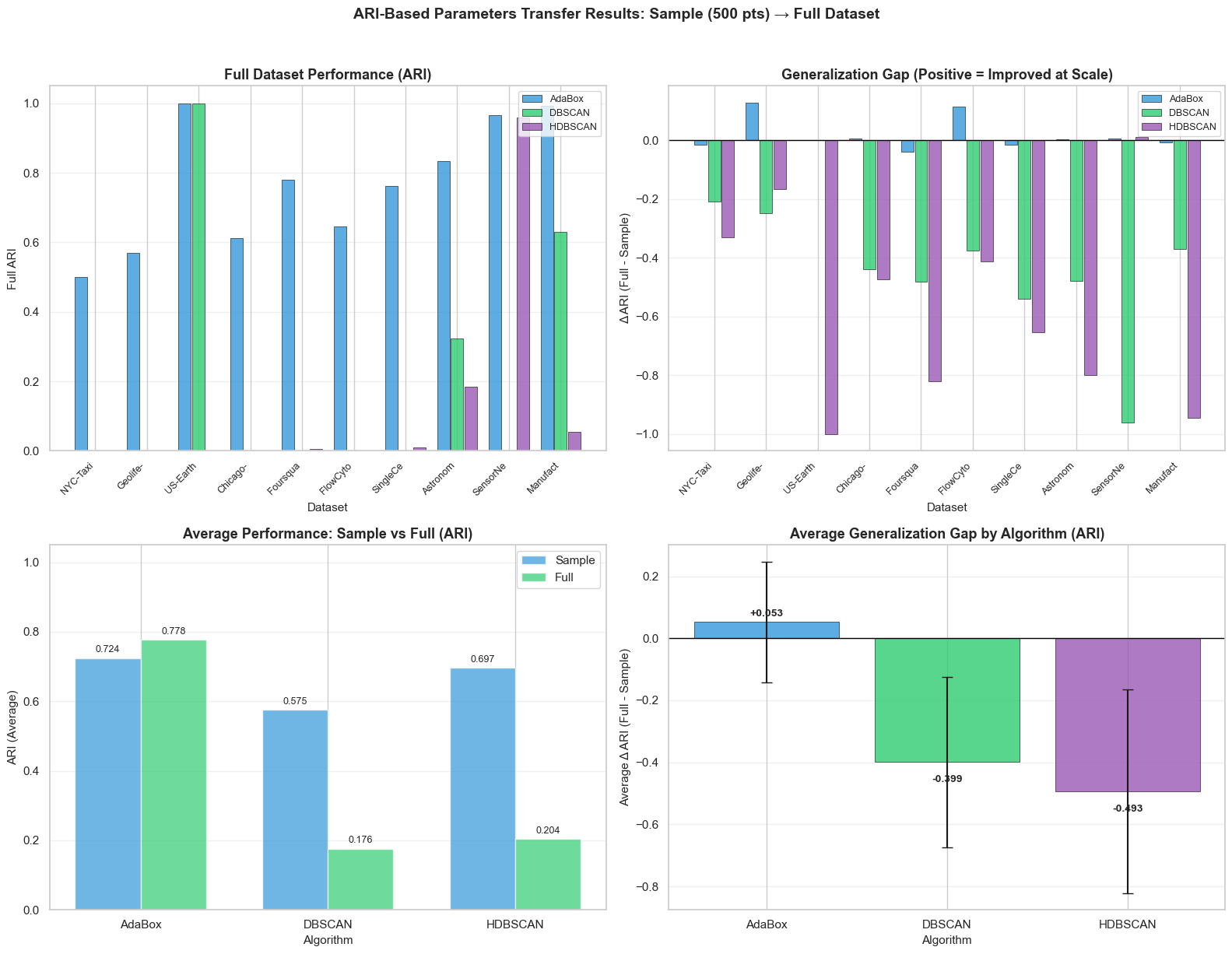}
\caption{Protocol~A Parameter Generalization Results. Four-panel visualization showing: (a)~Full dataset ARI across 10 datasets, (b)~$\Delta$ARI per dataset, (c)~Aggregate performance comparison, (d)~Average generalization gap with error bars.}
\label{fig:protA}
\end{figure}

Protocol~B reveals an even clearer pattern: baseline algorithms fail so early in the scaling ladder that subsequent stages become unnecessary.

\begin{table}[h]
\centering
\caption{Protocol~B Stage~2 results---parameter transfer to 100K points.}
\label{tab:protB}
\small
\begin{tabular}{lcccc}
\toprule
\textbf{Algorithm} & \textbf{Datasets Tested} & \textbf{Mean $\Delta$ARI} & \textbf{Mean ARI at 100K} & \textbf{Transfer Success} \\
\midrule
\textbf{AdaBox} & 10/10 & $-0.001$ & 0.733 & \textbf{10/10 (100\%)} \\
DBSCAN & 2/10 & $-0.257$ & 0.193 & 1/10 (10\%) \\
HDBSCAN & 4/10 & $-0.156$ & 0.522 & 2/10 (20\%) \\
\bottomrule
\end{tabular}
\end{table}

\begin{figure}[h]
\centering
\includegraphics[width=0.85\textwidth]{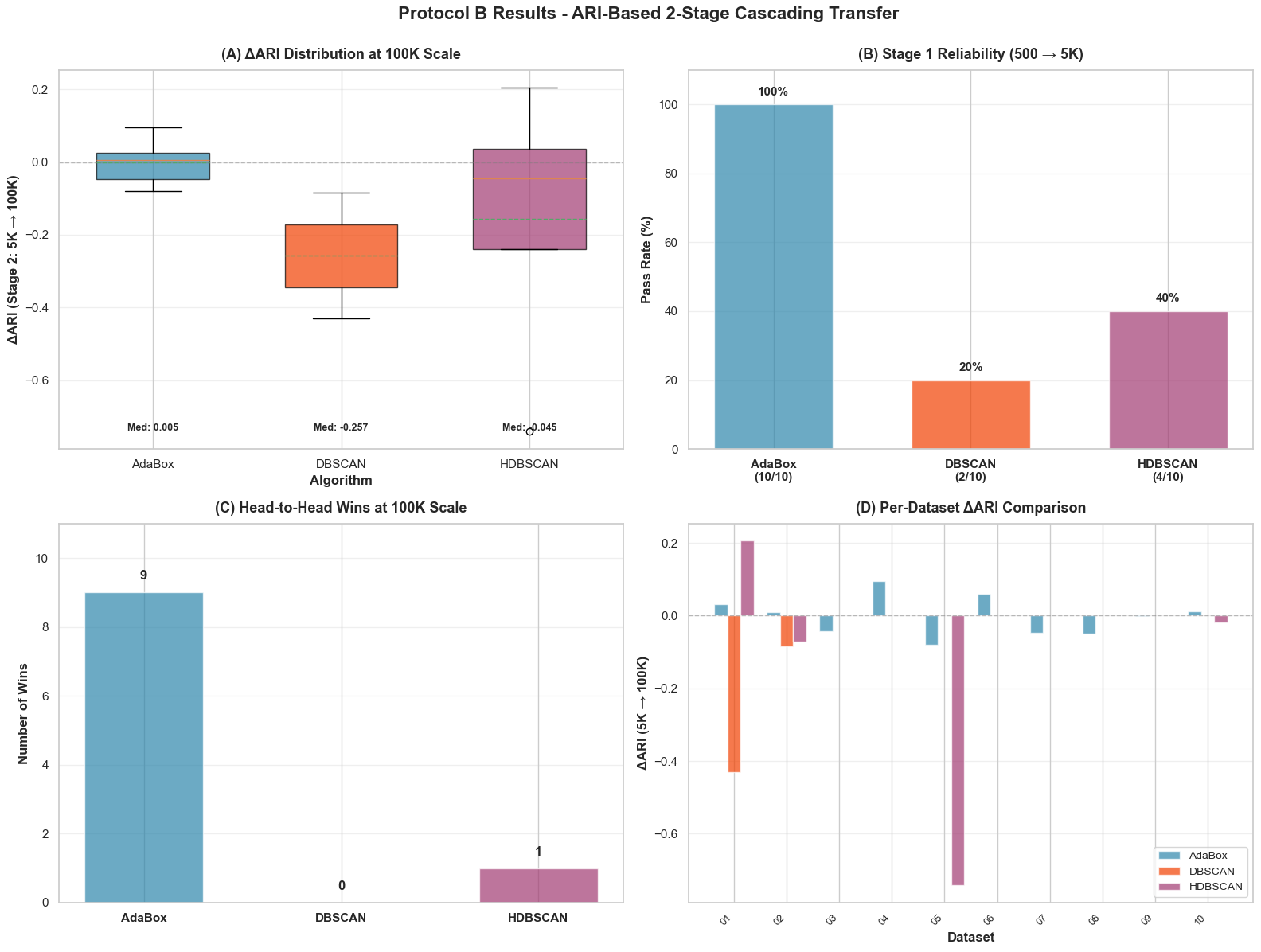}
\caption{Protocol~B Staged Transfer Results. Four-panel visualization showing: (a)~$\Delta$ARI distribution at 100K, (b)~Stage~1 pass rates, (c)~Head-to-head wins at 100K, (d)~Per-dataset $\Delta$ARI comparison.}
\label{fig:protB}
\end{figure}

Table~\ref{tab:protB} shows that, at the first checkpoint (500 $\rightarrow$ 5K, 10$\times$ scale), DBSCAN failed on 80\% of datasets and HDBSCAN failed on 60\%. Only AdaBox maintained 100\% pass rate. For algorithms surviving to Stage~2 (200$\times$ scale), AdaBox maintained near-perfect transfer with mean $\Delta$ARI = $-0.001$ and mean ARI = 0.733 at 100K scale. More details are shown in Figure~\ref{fig:protB}.

\subsection{RQ3: Ablation Study}

To validate that AdaBox's performance stems from principled design, we conduct ablation studies on 24 component-targeted datasets. We evaluate three ablation variants: No Merging, No Boundary Refinement, and Fixed Grid.

\begin{table}[h]
\centering
\caption{Ablation study results across 24 datasets.}
\label{tab:ablation}
\small
\begin{tabular}{lcccc}
\toprule
\textbf{Component Removed} & \textbf{Mean ARI} & $\boldsymbol{\Delta}$\textbf{ARI} & \textbf{\% Change} & $\boldsymbol{p}$\textbf{-value} \\
\midrule
Full AdaBox & 0.880 & --- & --- & --- \\
No Adaptive Merging & 0.695 & $-0.186$ & $-21.1\%$ & $< 0.0005$ \\
No Boundary Refinement & 0.800 & $-0.080$ & $-9.1\%$ & $< 0.0001$ \\
Fixed Grid & 0.716 & $-0.165$ & $-18.8\%$ & $< 0.0001$ \\
\bottomrule
\end{tabular}
\end{table}

The ablation reveals that all three major components contribute significantly: Adaptive Merging ($-21.1\%$), Adaptive Grid ($-18.8\%$), and Boundary Refinement ($-9.1\%$), all with $p < 0.001$.

\subsection{Summary of Findings}

\textbf{RQ1 (Performance):} AdaBox significantly outperforms DBSCAN and HDBSCAN across all five metrics ($p < 0.01$), achieving the best score on 78\% of datasets.

\textbf{RQ2 (Generalization):} AdaBox uniquely exhibits parameter generalization. In Protocol~A, AdaBox shows +7.3\% improvement while baselines degrade $\sim$70\%. In Protocol~B, AdaBox maintains 100\% pass rate with $\Delta$ARI = $-0.001$ at 200$\times$ scale, winning 9/10 head-to-head comparisons.

\textbf{RQ3 (Design):} Ablation confirms all components are statistically significant contributors.

\section{Conclusion}

We presented AdaBox, a density-based clustering algorithm that achieves state-of-the-art performance while exhibiting a unique parameter generalization property. Through a six-parameter design where five parameters capture scale-invariant cluster structure and one is systematically adjusted, AdaBox addresses the fundamental parameter sensitivity that limits practical deployment of existing methods.

Our comprehensive evaluation on 111 datasets yielded three key findings: (1)~Superior clustering quality with 78\% win rate; (2)~Robust parameter generalization with +7.3\% improvement at 30--100$\times$ scaling and 100\% reliability at 200$\times$ scaling where baselines fail; (3)~Validated component design with all ablation effects significant.

AdaBox's parameter generalization property has significant implications for large-scale clustering applications. Practitioners can now tune parameters on manageable samples with confidence that quality will be maintained at deployment scale, reducing the computational burden from $O(\text{full dataset})$ to $O(\text{sample size})$.

Future work includes formalizing the theoretical framework for transfer bounds, exploring online and streaming variants, and extending to higher-dimensional embedding spaces.

\bibliography{references}

\end{document}